\documentclass[conference]{IEEEtran}
\usepackage{times}

\usepackage[numbers]{natbib}

\usepackage{multicol}
\usepackage{url}
\usepackage{mathtools,fancyhdr,amsmath,amssymb,amsthm}
\usepackage{bm,upgreek}
\usepackage{algorithm,algpseudocode}

\newcommand{\argmax}{\operatornamewithlimits{arg\,max}}
\DeclarePairedDelimiter\abs{\lvert}{\rvert}
\DeclarePairedDelimiter\norm{\lVert}{\rVert}

\newcommand\blfootnote[1]{%
	\begingroup
	\renewcommand\thefootnote{}\footnote{#1}%
	\addtocounter{footnote}{-1}%
	\endgroup
}

\newcommand{\subparagraph}{}
\usepackage{titlesec}
\titlespacing*{\subsection}{0pt}{0.15\baselineskip}{0.05\baselineskip}
\titlespacing*{\section}{0pt}{0.6\baselineskip}{0.5\baselineskip}

\usepackage[keeplastbox]{flushend}

\newtheorem{remark}{Remark}

\newtheorem{problem}{Problem}

\begin{document}
\abovedisplayskip=5pt
\abovedisplayshortskip=5pt
\belowdisplayskip=5pt
\belowdisplayshortskip=5pt

\title{Active Preference-Based Gaussian Process Regression for Reward Learning}

\author{\authorblockN{Erdem B\i y\i k$^*$}
\authorblockA{Electrical Engineering\\
Stanford University\\
ebiyik@stanford.edu}
\and
\authorblockN{Nicolas Huynh$^*$}
\authorblockA{Applied Mathematics\\
École Polytechnique\\
nicolas.huynh@polytechnique.edu}
\and
\authorblockN{Mykel J. Kochenderfer}
\authorblockA{Aeronautics \& Astronautics\\
Stanford University\\
mykel@stanford.edu}
\and
\authorblockN{Dorsa Sadigh}
\authorblockA{Computer Science\\
Stanford University\\
dorsa@cs.stanford.edu}}

\maketitle

\begin{abstract}
\blfootnote{*First two authors contributed equally and are listed in alphabetical order.}
Designing reward functions is a challenging problem in AI and robotics. Humans usually have a difficult time directly specifying all the desirable behaviors that a robot needs to optimize. 
One common approach is to learn reward functions from collected expert demonstrations.
However, learning reward functions from demonstrations introduces many challenges: some methods require highly structured models, e.g. reward functions that are linear in some predefined set of features, while others adopt less structured reward functions that on the other hand require tremendous amount of data.
In addition, humans tend to have a difficult time providing demonstrations on robots with high degrees of freedom, or even quantifying reward values for given demonstrations.
To address these challenges, we present a preference-based learning approach, where as an alternative, the human feedback is only in the form of comparisons between trajectories. 
Furthermore, we do not assume highly constrained structures on the reward function. Instead,  we model the reward function using a Gaussian Process (GP) and propose a mathematical formulation to \emph{actively} find a GP using only human preferences. 
Our approach enables us to tackle both inflexibility and data-inefficiency problems within a preference-based learning framework. Our results in simulations and a user study suggest that our approach can efficiently learn expressive reward functions for robotics tasks. 
\end{abstract}

\IEEEpeerreviewmaketitle

\section{Introduction} \label{sec:intro}
Planning for robots that can act in a diverse set of environments based on human preferences can be quite challenging. It is generally infeasible for human designers to directly program the desired behavior for the full spectrum of possible situations. Hence, roboticists often use machine learning in at least part of their design to discover human preferences. One approach is to directly learn a robot policy using expert demonstrations \cite{ho2016generative,ross2013learning,song2018multi,stadie2017third}. However, in many interactive settings, we are interested in more generally learning a reward function that represents how a robot should act or interact in the world. 

Reward functions are powerful tools for specifying desirable robot behaviors, e.g. how to act safely, or what styles or goals the robot needs to follow. Unfortunately, specifying reward functions is not an easy task for human designers~\cite{clark2016faulty,ng1999policy,christiano2017deep}. 
Our goal in this work is to develop a \emph{data-efficient} method that can learn \emph{expressive} reward functions.

\begin{figure}[t]
	\includegraphics[width=\columnwidth]{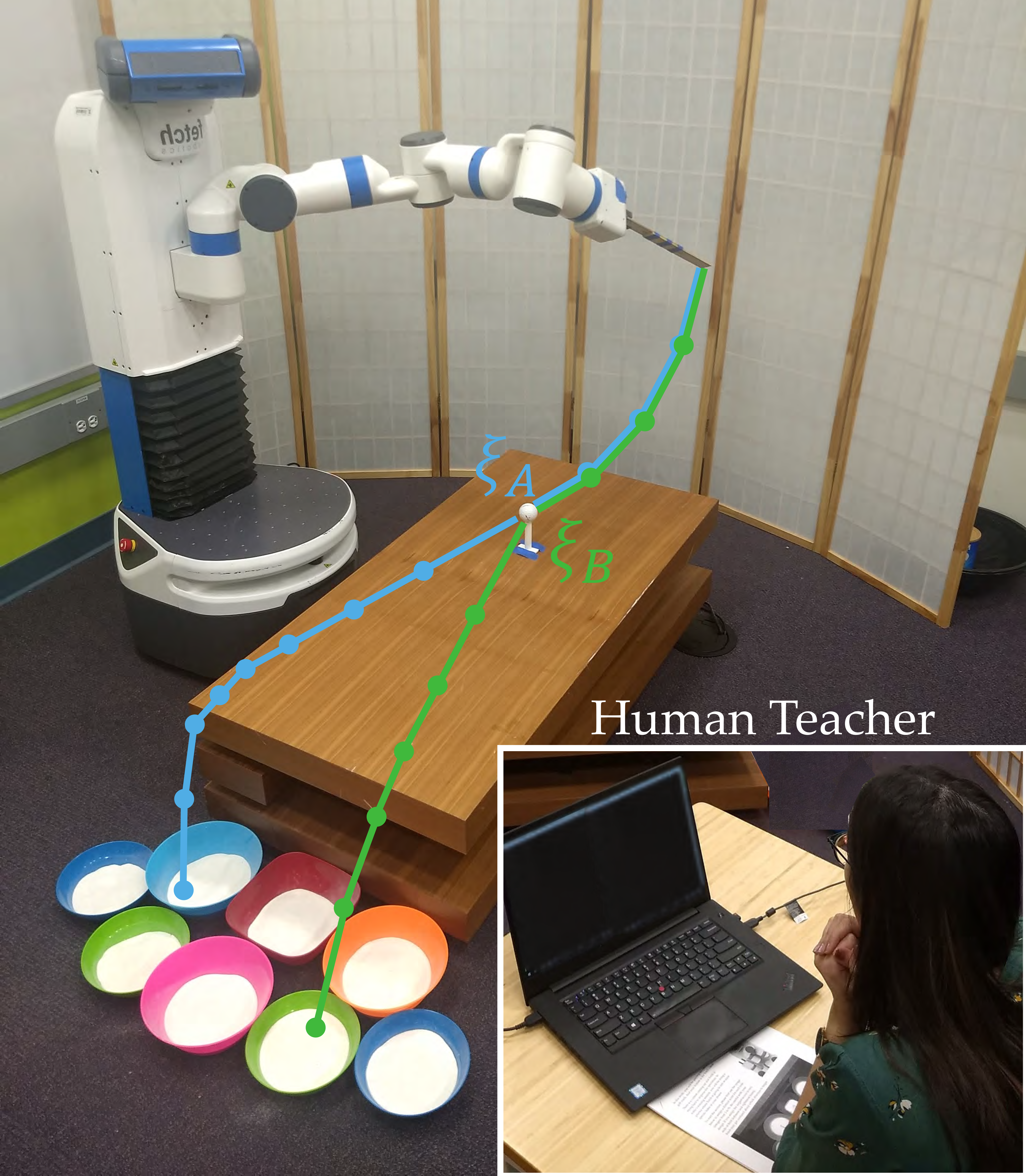}
	\centering
	\vspace{-15px}
	\caption{The user is trying to teach the robot how to play a variant of mini-golf, where the reward differs among eight targets. In preference-based learning, instead of trying to design a reward function by hand or controlling the robot to provide demonstrations, the user simply compares two demonstrated trajectories on the robot. Here, $\xi_A$ and $\xi_B$ demonstrate two trajectories that correspond to hitting the ball towards the blue or green targets.}
	\label{fig:front_fig}
	\vspace{-15px}
\end{figure}


Prior work has considered using a sequence of pairwise comparisons between trajectories as a successful technique to learn reward functions~\cite{cakmak2011human,ibarz2018reward,brown2019deep,tucker2019preference,sadigh2017active,wilde2019bayesian,akrour2012april,lepird2015bayesian}. For example, as shown in Fig.~\ref{fig:front_fig}, the robot can demonstrate the blue and green trajectories, $\xi_A$ and $\xi_B$, and ask the human designer to compare the two. Preference-based reward learning can then leverage a sequence of pairwise comparisons to accurately estimate a reward function.


However, preference-based learning techniques are in general not very data-efficient, as each pairwise comparison only provides $1$ bit of information, i.e., if $\xi_A$ is preferred over $\xi_B$ or vice versa. 
Therefore, active learning is commonly employed in this framework to find the most informative or diverse sequence of questions for efficiently converging to the underlying reward function~\cite{sadigh2017active,wilde2019bayesian,akrour2012april,biyik2018batch,biyik2019batch,biyik2019asking,basu2019active,palan2019learning,katz2019learning,biyik2019green,racca2019teacher}.

Unfortunately, most prior active reward learning works rely on a strong assumption about the structure of the reward function, i.e., \emph{the reward function is a linear combination of a set of hand-coded features}. While this assumption is commonly needed for active learning to scale, it is very limiting because linear reward functions are not sufficiently expressive. For example, a linear reward would require several features for the human teacher to be able to specify every reward configuration of targets in the task demonstrated in Fig.~\ref{fig:front_fig}, i.e., how targets compare to each other. The features to this task could be, for example, distances to each and every target. On the other hand, if the reward model was nonlinear, one can capture all possible configurations with only two features: speed for how far the ball will be thrown, and angle for which direction to shoot. While neural networks or kernel functions can provide this flexibility, these techniques considerably increase the number of parameters needed, which prohibits (or renders useless the advantage of) active learning algorithms.
\begin{quote}
    Our key insight is to model the reward function using a Gaussian Process (GP) \cite{rasmussen2005gaussian}. GPs are non-parametric models that can capture nonlinearities, allowing us to actively and efficiently learn highly expressive reward functions.
\end{quote}

In this work, we propose a mathematical framework for \emph{actively} fitting a GP using only pairwise comparisons between two trajectories, which we call \emph{preference} data. Leveraging GPs, instead of linear models with hand-designed features, improves the expressiveness of reward functions by incorporating nontrivial nonlinearities. Besides, our active query generation method enables us to still get the benefits of active learning.

We make two main contributions in this work:
\begin{itemize}
	\item We propose a \emph{data-efficient} mathematical framework for actively regressing a GP with preference data to learn \emph{expressive} reward functions from humans.
	\item We demonstrate the performance of our framework through simulated environments and user studies on a manipulator robot playing a variant of mini-golf based on different human preferences. Our results show our approach can be used for reward learning in complex and interesting settings and is more data-efficient than various other baselines.
\end{itemize}

\section{Related Work}
In this section, we will discuss the prior work focused on learning reward functions from demonstration, or other sources of data, as well as related work in Gaussian process regression and its relation to our work.

\vspace{5pt}

\noindent\textbf{Learning reward functions from demonstrations.} Prior work has studied learning reward functions from collected expert demonstrations. This is commonly referred to as inverse reinforcement learning (IRL), where we assume human demonstrations are based on them approximately optimizing a reward function~\cite{abbeel2004apprenticeship,abbeel2005exploration,ng2000algorithms,ramachandran2007bayesian,ziebart2008maximum}. The learned reward function can then be used by a robot to optimize its actions in the broad range of environments.


While IRL has shown promising results in a variety of domains, robots, especially manipulators with high degrees of freedom, are often too difficult to manually operate~\cite{losey2020controlling,akgun2012keyframe,dragan2012formalizing,javdani2015shared,khurshid2015data}. Moreover, recent studies in autonomous driving, where the high degrees of freedom of a robot is not an issue, suggest that people do not prefer an autonomous car to follow their own demonstrations and instead prefer a different behavior that tends to be more timid~\cite{basu2017you}. These indicate that one needs to go beyond human demonstrations to properly capture the preferred reward function.

In our framework, instead of relying on human demonstrations, we learn the reward functions through the preference queries. Therefore, our method does not require experts who can control the system in the desirable way. 

\vspace{5pt}

\noindent\textbf{Learning reward functions from other sources of data.}
In addition to demonstrations and physical corrections \cite{bajcsy2017learning,bajcsy2018learning}, where the robot attempts to learn the reward function through physical human interference, learning from rankings \cite{brown2019deep} is another popular approach. A particular case of this is when the rankings are only pairwise comparisons, which we call preference queries. 
Previous works have investigated the use of preference queries for reward learning. \citet{sadigh2017active} proposed an acquisition function to actively generate the queries. Further studies extended this approach to batch-active methods \cite{biyik2018batch,biyik2019batch}, rankings instead of pairwise comparisons \cite{biyik2019green}, general Markov Decision Process (MDP) settings \cite{katz2019learning}, and settings that integrate expert demonstrations with preference queries \cite{palan2019learning}. The reward function these prior works assume is linear with respect to some hand-coded features. This assumption limits the model flexibility and requires very careful feature design. \citet{basu2019active} explored modeling a multi-modal reward function, but the reward was still linear in each mode. Furthermore, they focused only on bi-modal distributions. Scalability to more modes remains an issue.

In this work, we do not make the linearity assumption and instead model the reward using a GP. Our results show this significantly improves the expressive power of the learned reward function, and the method is still very data-efficient.

\vspace{5pt}
\noindent\textbf{Gaussian process regression.} On the machine learning side, \citet{gonzalez2017preferential} and \citet{chu2005preference} proposed methods for preference-based Bayesian optimization and GP regression, respectively, but they were not active. Furthermore, \cite{gonzalez2017preferential} required to regress a GP over $2d$-dimensions to model a $d$-dimensional function, which causes a computational burden. More relevantly, \citet{houlsby2012collaborative} presented an active method for preference-based GP regression. However, similar to \cite{gonzalez2017preferential}, the regression was over a $2d$-dimensional space where the learned model predicts the outcome of a comparison rather than outputting a reward value. \citet{jensen2011pairwise} showed how to update a GP with preference data, but the active query generation was not an interest. \citet{kapoor2007active} developed an active learning approach for classification with GPs. This is a specific case of our problem, as the labels in classification are consistent, i.e., the labels assigned to the samples in the dataset, even if they are incorrect, do not change during training. In our case, the user can respond to the same preference query inconsistently depending on their noise model. \citet{houlsby2011bayesian} and \citet{daniel2015active} proposed active GP fitting methods for classification and reward learning, respectively. While the latter focused on robotics tasks, they were not preference-based. Hence, they may be infeasible in many applications as it is difficult for humans to assign actual reward values. 

In this work, we propose an active query generation method for preference-based GP regression. While being data-efficient, this method also does not require the humans to assign actual reward values to the trajectories for fitting the GP.


\section{Problem Formulation} \label{sec:problem}
We model the environment the robot is going to operate in as a finite-horizon deterministic MDP. We use $s_t\in\mathcal{S}$ to denote the state and $a_t\in\mathcal{A}$ for the action (control inputs) at time $t$. A trajectory $\xi\in\Xi$ within this MDP is a sequence that consists of the initial position and the actions: $\xi = (s_0,a_0,a_1,\dots,a_T)$, where $T$ is the finite time horizon. 

We assume a reward function over trajectories, $R:\Xi\to\mathbb{R}$, that encodes the human user's preferences about how they want the system to operate.

We assume the reward function $R$ can be formulated as $R(\xi) = f(\Psi(\xi))$, where $\Psi:\Xi\to\mathbb{R}^d$ defines a set of trajectory features, e.g. average speed and minimum distance to the closest car in a driving trajectory. However, we emphasize that this formulation of $R$ enables a more general form of functions that does not require strong assumptions -- such as linearity in the features -- which is commonly put in place when learning reward functions. We use a GP to model $f$, which allows us to avoid strong assumptions about the form of $f$.\footnote{Due to computation reasons, we assume $d$ is small. Compared to previous works which assume $R$ to be linear in features, this is a very mild assumption.}

Our goal is to learn this more general form of reward functions using preference data in the form of pairwise comparisons. The robot will demonstrate a query $Q$ consisting of two trajectories, $\xi_A$ and $\xi_B$ as shown in Fig.~\ref{fig:front_fig} with blue and green curves, to the user, and will ask which trajectory they prefer. The user will respond to this query based on their preferences. The user's response provides useful information about the underlying preference reward function $R$. Of course, we cannot assume human responses are perfect for every query, so we model the noise in their responses using the commonly adopted probit model, which assumes humans make a binary decision between the two trajectories noisily based on the cumulative distribution function (cdf) of the difference between the two reward values:
\begin{align*}
P(q=\xi_A \mid Q=\{\xi_A,\xi_B\}) = P\left(R(\xi_A) - R(\xi_B) > v\right),
\end{align*}
where $q\in Q$ denotes the user's choice, and $v\sim\mathcal{N}(0,2\sigma^2)$ for some standard deviation $\sigma\sqrt{2}$. Therefore, equivalently:
\begin{align}
P(q=\xi_A \mid Q=\{\xi_A,\xi_B\}) = \Phi\left(\frac{R(\xi_A)-R(\xi_B)}{\sqrt{2}\sigma}\right),
\label{eq:human_model}
\end{align}
where $\Phi$ is the cumulative distribution function of the standard normal.

Having defined the problem setting, we are now ready to present our method for learning data-efficient and expressive reward functions using GPs.

\section{Methods} \label{sec:methods}
In this section, we first give some background information about Gaussian Processes. We then introduce preference-based GP regression, where we show how to update a GP with the results of pairwise comparisons. Finally, we present our approach to active preference query generation, where we discuss how to find the most informative query that accelerates the learning.\footnote{We make our Python code for active query generation publicly available at \url{https://github.com/Stanford-ILIAD/active-preference-based-gpr}.} To simplify the notation, we replace $\Psi(\xi)$ with $\Psi$, with superscripts and subscripts when needed.

\vspace{5px}
\subsection{Gaussian Processes}
We start by introducing the necessary background on GPs for our work. We refer the readers to \cite{rasmussen2005gaussian} for other uses of GPs in machine learning.

Suppose we are given a dataset $\bm{\Psi}\!=\!(\Psi_i)_{i=1}^N$, where $\Psi_i\!\in\!\mathbb{R}^d$. We employ a probabilistic point of view for $f$ by modeling it using a GP, which enables us to model nonlinearities and uncertainties well without introducing parameters. We have:
\begin{align}
P(\mathbf{f} \mid \bm{\upmu}, \mathbf{K}) = \frac{\exp\left(-\frac12 (\mathbf{f}-\bm{\upmu})^\top \mathbf{K}^{-1}(\mathbf{f}-\bm{\upmu})\right)}{(2\pi)^{N/2} \abs{\mathbf{K}}^{1/2}} ,
\label{eq:gp_prob}
\end{align}
where $\mathbf{f}=(f(\Psi_i))_{i=1}^N$, $\bm{\upmu}$ and $\mathbf{K}$ are the mean vector and the covariance matrix of the GP distribution for the $N$ items in the dataset. Put it in another way, $\mathbf{f}$ follows a multivariate distribution. The covariance matrix depends on the used kernel. In this work, we use a variant of radial basis function (RBF) kernel with hyperparameter $\theta$:
\begin{align*}
k(\Psi_i,\Psi_j) &= \exp\left(-\theta\norm{\Psi_i-\Psi_j}_2^2\right) - \bar{k}(\Psi_i,\Psi_j),\\
\bar{k}(\Psi_i,\Psi_j) &= \exp\left(-\theta\norm{\Psi_i-\bar{\Psi}}_2^2 - \theta\norm{\Psi_j-\bar{\Psi}}_2^2\right),
\end{align*}
where $\bar{\Psi}\in\mathbb{R}^d$ is an arbitrary point for which we assume $f(\bar{\Psi})=0$. This is important because the query responses only depend on the relative difference between the two reward function values at the trajectories, i.e., $f(\Psi)+c$ for any $c\in\mathbb{R}$ would have the same likelihood for a dataset as $f(\Psi)$. By setting $f(\bar{\Psi})=0$ for some arbitrary $\bar{\Psi}\in\mathbb{R}^d$, we dissolve this ambiguity. It does not introduce an assumption because for any function $f'$ and for any point $\bar{\Psi}$, one can define $f(\Psi) = f'(\Psi) - f'(\bar{\Psi})$ without loss of generality\textemdash both $f'$ and $f$ will encode the same preferences. Finally, this variant of the RBF kernel is still positive semi-definite, because it is equivalent to the covariance kernel of a GP which is initialized with an initial data point and a standard RBF kernel prior.

\vspace{5px}
\subsection{Preference-based GP Regression}
In preference-based learning, we have a dataset $\bm{\Psi}\!=\!((\Psi_i^{(1)},\Psi_i^{(2)}))_{i=1}^N$, consisting of pairs of trajectories $\Psi_i^{(1)},\!\Psi_i^{(2)}\!\in\!\mathbb{R}^d$, and user responses $\mathbf{q}\!=\!(q_i)_{i=1}^N$, where $q_i\!\in\!\{0,1\}$ indicates whether the user preferred $\Psi_i^{(1)}$ or $\Psi_i^{(2)}$. Accordingly, $\mathbf{K}$ is now a $2N\!\times\!2N$ matrix, whose rows and columns correspond to $\Psi_i^j, \forall i\!\in\!\{1,\dots,N\}, \forall j\!\in\!\{1,2\}$. Similarly, $\bm{\upmu}$ is a $2N$-vector. Using a Bayesian approach to update the GP with new preference data $(\Psi,q)$ gives:
\begin{align}
P(f \mid \bm{\upmu}, \mathbf{K}, \Psi, q) &\propto P(q \mid f, \bm{\upmu}, \mathbf{K}, \Psi)P(f \mid \bm{\upmu},\mathbf{K}, \Psi) \nonumber\\
&= P(q \mid f, \Psi)P(f \mid \bm{\upmu},\mathbf{K}).
\label{eq:bayesian_update}
\end{align}
Here, the first term is just the probabilistic human response model given in Eqn.~\eqref{eq:human_model}, and the second term is Eqn.~\eqref{eq:gp_prob}. However, this posterior does not follow a GP distribution similar to Eqn.~\eqref{eq:gp_prob}, and becomes analytically intractable \cite{jensen2011pairwise}.

Prior works have shown it is possible to perform some approximation such that the posterior is another GP \cite{jensen2011pairwise,rasmussen2005gaussian}. The most common approximation techniques are  
\begin{itemize}
    \item Laplace approximation, where the idea is to model the posterior as a GP such that the mode of the likelihood is treated as the posterior mean, and a second-order Taylor approximation around the maximum of the log-likelihood gives the posterior covariance. This technique is computationally very fast. 
    \item Expectation Propagation (EP), where the idea is to approximate each factor of the product by a Gaussian. EP is an iterative method that processes each factor iteratively to update the distribution to minimize its KL-divergence with the true posterior. While it is more accurate than Laplace approximation, it is slower in practice \cite{nickisch2008approximations}.
\end{itemize} 
In this paper, we use the former for its computational efficiency. Hence, we now show how to compute the quantities for Laplace approximation, i.e., posterior mean and covariance.

\vspace{5pt}
\noindent\textbf{Finding the posterior mean.} We use the mode of the posterior as an approximation to the posterior mean:
\begin{align}
\argmax_{\mathbf{f}} \left(\log\left(p(\mathbf{q}\mid \bm{\Psi}, \mathbf{f})\right) + \log\left(P(\mathbf{f}\mid \bm{\Psi})\right)\right)
\label{eq:posterior_mean}
\end{align}
Because the preference data are conditionally independent, the first term can be written as:
\begin{align*}
\log\left(P(\mathbf{q}\mid \bm{\Psi}, \mathbf{f})\right) &= \sum_{i=1}^N{\log P(q_i \mid \Psi_i, \mathbf{f})}\\
&= \sum_{i=1}^N{\log \Phi\left(\frac{f(\Psi_i^{(1)})-f(\Psi_i^{(2)})}{\sqrt{2}\sigma}\right)}
\end{align*}
due to Eqn.~\eqref{eq:human_model}. Adopting a zero-mean prior for $f$, we can write the second term of the optimization~\eqref{eq:posterior_mean} as:
\begin{align*}
\log\left(P(\mathbf{f}\mid \bm{\Psi})\right) &= -\frac12 \log\abs{\mathbf{K}} - N\log{2\pi} - \frac12 \mathbf{f}^\top \mathbf{K}^{-1}\mathbf{f}
\end{align*}
Armed with these two expressions, we can now optimize the log-likelihood and thus find the mode of it to approximate the posterior mean.

\vspace{5pt}
\noindent\textbf{Finding the posterior covariance matrix.} Following \cite{rasmussen2005gaussian}, and omitting the derivation details for brevity, the posterior covariance matrix is $(\mathbf{K}^{-1} + \mathbf{W})^{-1}$ where $\mathbf{W}$ is the negative Hessian of the log-likelihood:
\begin{align*}
W_{ij} = -\frac{\partial^2 \log\left(P(\mathbf{q}\mid \bm{\Psi}, \mathbf{f})\right)}{\partial f^{(i)} \partial f^{(j)}}.
\end{align*}

Now, we know how to approximate the posterior mean and the posterior covariance for the Laplace approximation of Eqn.~\eqref{eq:bayesian_update}. This allows us to model and update the reward with preference data using a GP.

We also want to perform inference from this approximated GP. Inference is not only useful for active query generation, but it also enables us to compute the reward expectations and variances given a trajectory.

\vspace{5pt}
\noindent\textbf{Inference.} Given two points $\Psi^{(1)}_*, \Psi^{(2)}_* \in \mathbb{R}^d$, we want to be able to compute the expected mean rewards $\bm{\upmu}_*$ and also the covariance matrix between those two points $\mathbf{K}_*$, both of which will be useful for active query generation. These are given by:
\begin{align}
\mu_* = \mathbb{E}\left[\mathbf{f}_* \mid \bm{\Psi}, \mathbf{q}, \Psi^{(1)}_*, \Psi^{(2)}_*\right] = k_*^\top \mathbf{K}^{-1}\mathbf{f},
\label{eq:inference_mean}
\end{align}
where ${k_*}_{ij} = k(\Psi_*^{(i)}, \Psi_j)$ is a $2\times 2N$ matrix, and
\begin{align}
\mathbf{K}_* = \bm{K_0} - k_*\left(\bm{I}_{2N} + \mathbf{W}\mathbf{K}\right)^{-1}\mathbf{W}k_*^\top,
\label{eq:inference_cov}
\end{align}
where ${K_0}_{ij} = k\left(\Psi_*^{(i)},\Psi_*^{(j)}\right)$ is a $2\times 2$ matrix, $\bm{I}_{2N}$ is the $2N\times 2N$ identity matrix.

Equipped with all these tools which enable us to approximate the posterior distribution with a GP and perform inference over it, we are now ready to present our contributions on the active query generation.

\subsection{Active Query Synthesis for Reward Learning} \label{sec:active_reward_learning}
While we now know how to learn the reward function $f$ using only pairwise comparisons, this endeavor can require tremendous amount of data, because each query will give at most $1$ bit of information. Furthermore, we can expect a decreasing trend in the information gain as we learn the reward function. Therefore, it is important to select the queries for the human such that each query gives as much information as possible. For linear reward models, previous work has shown that this can be done by maximizing the mutual information, which also makes the queries easy for the user \cite{biyik2019asking}. Extending this formulation to the reward functions modeled with a GP is not trivial, because one needs to sample from the GP many times for each trajectory, whereas a linear reward form allows the reward prediction after sampling the linear weight terms only once.

Hence, for the active query generation, our goal is to perform information gain maximization with GPs.

\begin{problem}
Formally, we want to solve the following problem:
\begin{align*}
\Psi_*^{(1)}, \Psi_*^{(2)} = \argmax_{\Psi^{(1)}, \Psi^{(2)}} I(f;q \mid \Psi, \bm{\Psi}, \mathbf{q}),
\end{align*}
where $I$ is the mutual information and $q$ is the response to the query $\Psi=(\Psi^{(1)}, \Psi^{(2)})$. This optimization is equivalent to
\begin{align}
\argmax_{\Psi^{(1)}, \Psi^{(2)}} \left(H(q \mid \Psi, \bm{\Psi}, \mathbf{q}) - \mathbb{E}_{f\sim P(f \mid \bm{\Psi}, \mathbf{q})}\left[H(q \mid \Psi, f)\right]\right),
\label{eq:problem}
\end{align}
where $H$ is the information entropy.
\end{problem}

This optimization can be interpreted as follows:
On one hand, maximizing the first entropy term $H(q \mid \Psi, \bm{\Psi}, \mathbf{q})$ encourages fast convergence by maximizing the uncertainty of the outcome of every query for the learned GP model. On the other hand, minimizing the second entropy term $H(q \mid \Psi, f)$ encourages the ease of responding to the queries by the user meaning the user should be certain about their choices.


We defer the full derivation of \eqref{eq:problem} to the appendix, but here we give an easy-to-implement formulation of the optimization. Denoting the posterior mean of $f(\Psi^{(i)})$, which is obtained using Eqn.~\eqref{eq:inference_mean}, with $\mu^{(i)}$, the objective function can be written as
\begin{align}
h\left(\Phi\left(\frac{\mu^{(1)}-\mu^{(2)}}{\sqrt{2\sigma^2 + g(\Psi^{(1)},\Psi^{(2)})}}\right)\right) - m\left(\Psi\right)
\label{eq:active_query_generation}
\end{align}
where
\begin{align*}
g(\Psi^{(1)},\Psi^{(2)})=&\mathrm{Var}\left(f(\Psi^{(1)})\right) + \mathrm{Var}\left(f(\Psi^{(2)})\right)\\
& - 2 \: \mathrm{Cov}\left(f(\Psi^{(1)}),f(\Psi^{(2)})\right),
\end{align*}
whose terms can be computed using Eqn.~\eqref{eq:inference_cov};
$h$ is the binary entropy function, i.e., 
\begin{equation*}
    h(p) = -p\log_2(p) - (1-p)\log_2(1-p),
\end{equation*}
and
\begin{align*}
m\left(\Psi\right) = \frac{\sqrt{\pi\ln(2)\sigma^2}\exp\left(-\frac{(\mu^{(1)} - \mu^{(2)})^2}{\pi\ln(2)\sigma^2 + 2g(\Psi^{(1)},\Psi^{(2)})}\right)}{\sqrt{\pi\ln(2)\sigma^2 + 2g(\Psi^{(1)},\Psi^{(2)})}}.
\end{align*}

Synthesizing queries that maximize this objective will give us very informative data points for preference-based GP regression and improve data-efficiency.

Previously, \citet{biyik2019asking} have shown for the linear reward models that using an information gain based formulation accelerates the learning whereas volume removal based methods (such as \cite{sadigh2017active}) rely on local optima and can produce trivial queries that compare the exact same trajectory and so gives no information. In the following, we show our formulation also does not suffer from this trivial query problem.

\begin{remark}
The trivial query $\Psi= \{\Psi^{(A)},\Psi^{(A)}\}$ does not maximize our acquisition function given in \eqref{eq:active_query_generation}, and is in fact a global minimizer.
\end{remark}
\begin{proof}
For the query $\Psi\!=\!\{\Psi^{(A)},\!\Psi^{(A)}\}$, we rewrite \eqref{eq:active_query_generation} as 
\begin{align*}
    h\left(\Phi\left(\frac{\mu^{(A)}-\mu^{(A)}}{\sqrt{2\sigma^2 + g(\Psi^{(A)},\Psi^{(A)})}}\right)\right) - m\left(\Psi\right) &= 1 - m(\Psi)
\end{align*}
where $\mathrm{Var}\left(f(\Psi^{(A)})\right)=\mathrm{Cov}\left(f(\Psi^{(A)}),f(\Psi^{(A)})\right)$, and so $g(\Psi^{(A)},\Psi^{(A)})=0$, and
\begin{align*}
    m\left(\Psi\right) = \frac{\sqrt{\pi\ln(2)\sigma^2}\exp\left(-\frac{(\mu^{(A)} - \mu^{(A)})^2}{\pi\ln(2)\sigma^2 + 2g(\Psi^{(A)},\Psi^{(A)})}\right)}{\sqrt{\pi\ln(2)\sigma^2 + 2g(\Psi^{(A)},\Psi^{(A)})}}
    = 1
\end{align*}
which makes the objective value $0$. Since the information gain has to be nonnegative, this completes the proof that the trivial query is a global minimizer of the objective.
\end{proof}

\section{Simulation Experiments} \label{sec:simulation_experiments}
\begin{figure}[t]
	\includegraphics[width=0.9\columnwidth]{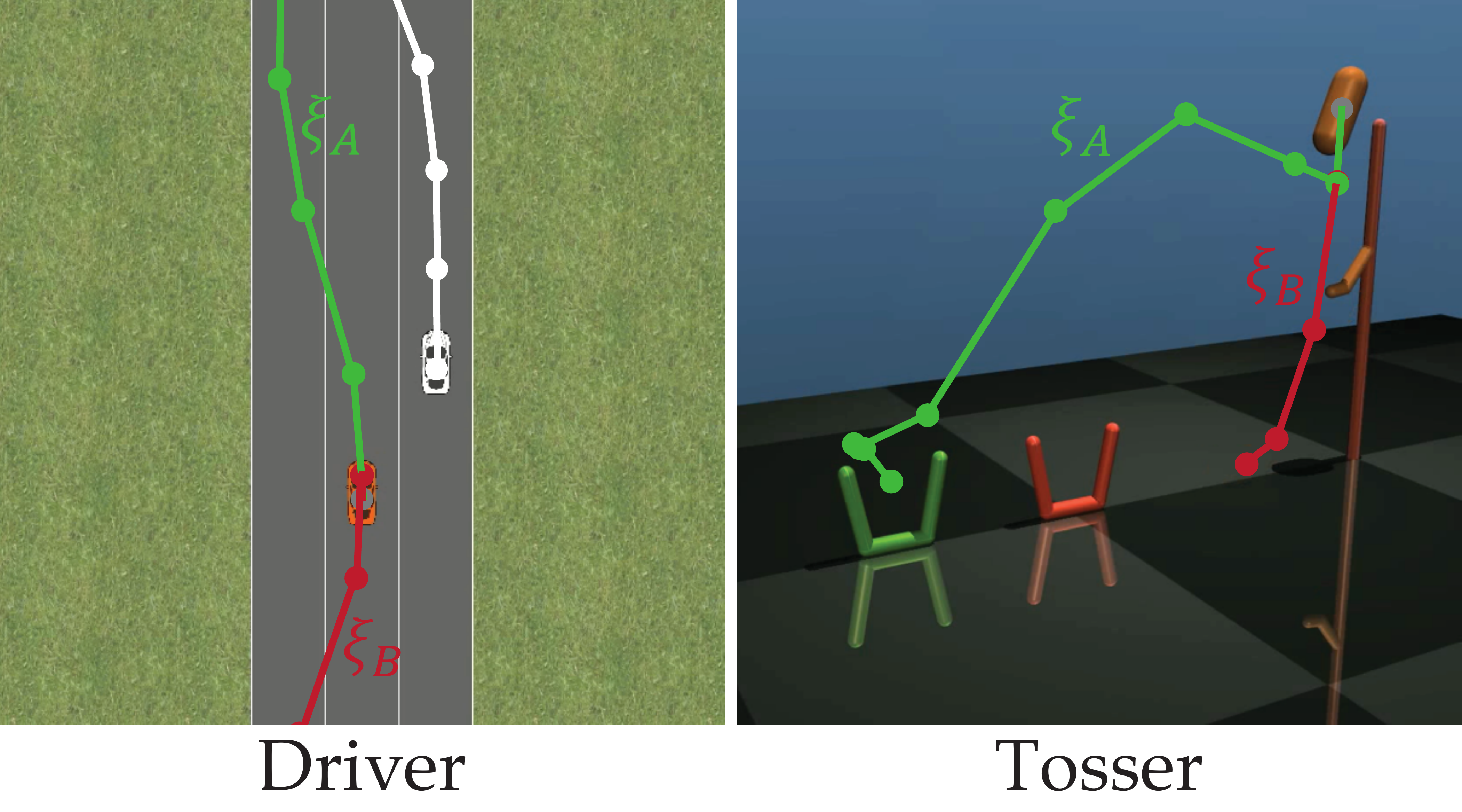}
	\centering
	\vspace{-10px}
	\caption{Sample trajectories are shown for the two simulation environments. In \emph{Driver}, another car is cutting in front of the ego vehicle. In \emph{Tosser}, the robot must hit the dropping capsule such that it will fall into one of the baskets.
	}
	\vspace{-15px}
	\label{fig:simulation_visuals}
\end{figure}
In this section, we present our experiments in two simulation domains to demonstrate how (i) GP rewards improve expressiveness over linear reward functions, and (ii) active query generation improves data-efficiency over random querying.


\vspace{5px}
\subsection{Simulation Environments}
To validate our framework on robotics tasks, we used two simulation environments: a 2D \emph{Driver} simulation \cite{sadigh2016planning} and a MuJoCo \cite{todorov2012mujoco} environment to simulate a \emph{Tosser} robot that tries to throw an object into a basket \cite{biyik2018batch}. We show visuals from these environments with sample trajectories in Fig.~\ref{fig:simulation_visuals}. For example in \emph{Driver}, the user is asked whether they would move forward or backward in the given scenario. While the users would have a common response to this query, some questions may differ among the users. For instance in Tosser, the query asks the user whether to throw the ball into the green basket or to drop it instead. Depending on the users' preferences about the green basket, different users may have different responses.

\begin{figure*}[h]
	\includegraphics[width=\textwidth]{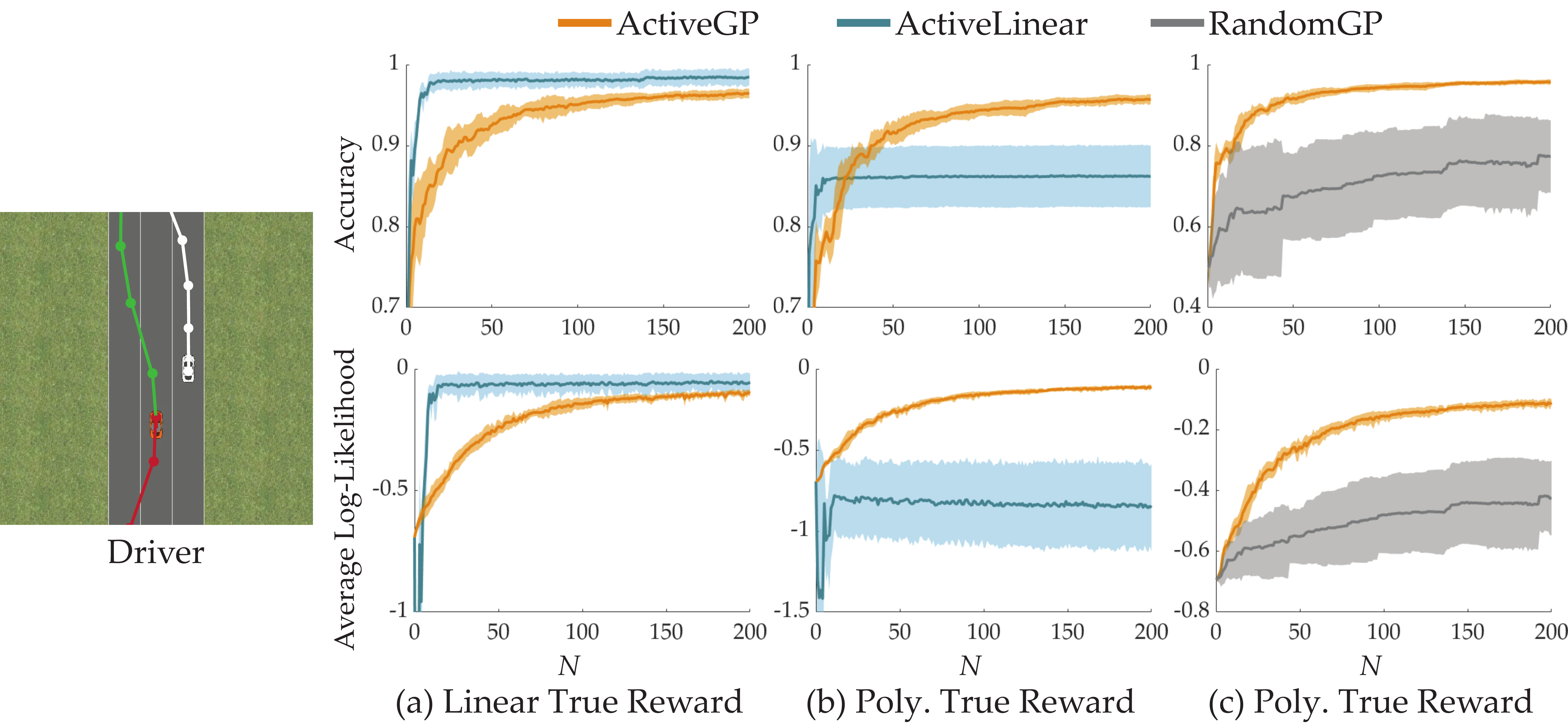}
	\centering
	\vspace{-20px}
	\caption{Accuracies and average log-likelihoods for test set queries are shown for the \emph{Driver} environment (mean$\pm$std over $5$ runs). \textbf{(a)} Expressiveness results when the true underlying reward function is linear. \textbf{(b)} Expressiveness results when the true underlying reward function is a degree-of-two polynomial. \textbf{(c)} Data-efficiency results that compare \textsc{ActiveGP} with \textsc{RandomGP}. Accuracies and average log-likelihoods for test set queries are shown (mean$\pm$std). Active query generation improves data-efficiency over random querying in both tasks. This can be seen through both accuracy and log-likelihood.}
	\label{fig:driver_results}
	\vspace{-15px}
\end{figure*}

In these two environments, we use the following simple features for the function $\Psi$ similar to \cite{biyik2018batch}:
\begin{itemize}
	\item \emph{Driver}: Distance to the other car, speed, heading angle, distance to the closest lane center.
	\item \emph{Tosser}: The maximum horizontal range, and the number of capsule flips.
\end{itemize}
In contrast to what the previous work reported, here we do not need to fine-tune the feature parameters to learn the reward functions because GPs can effectively capture nonlinearities.

\vspace{5pt}
\noindent \textbf{Simulated Human Model.} We simulated human responses with an underlying true reward function $f$ with some Gaussian noise, in accordance with Eqn.~\eqref{eq:human_model}. We modeled $f$ as either a degree-of-two polynomial or a linear function. In both cases, we selected the parameters of $f$ as i.i.d. random samples from the standard normal distribution. We repeated each simulation experiment $5$ times with varying underlying true reward functions.
\begin{figure}[h]
	\includegraphics[width=0.84\columnwidth]{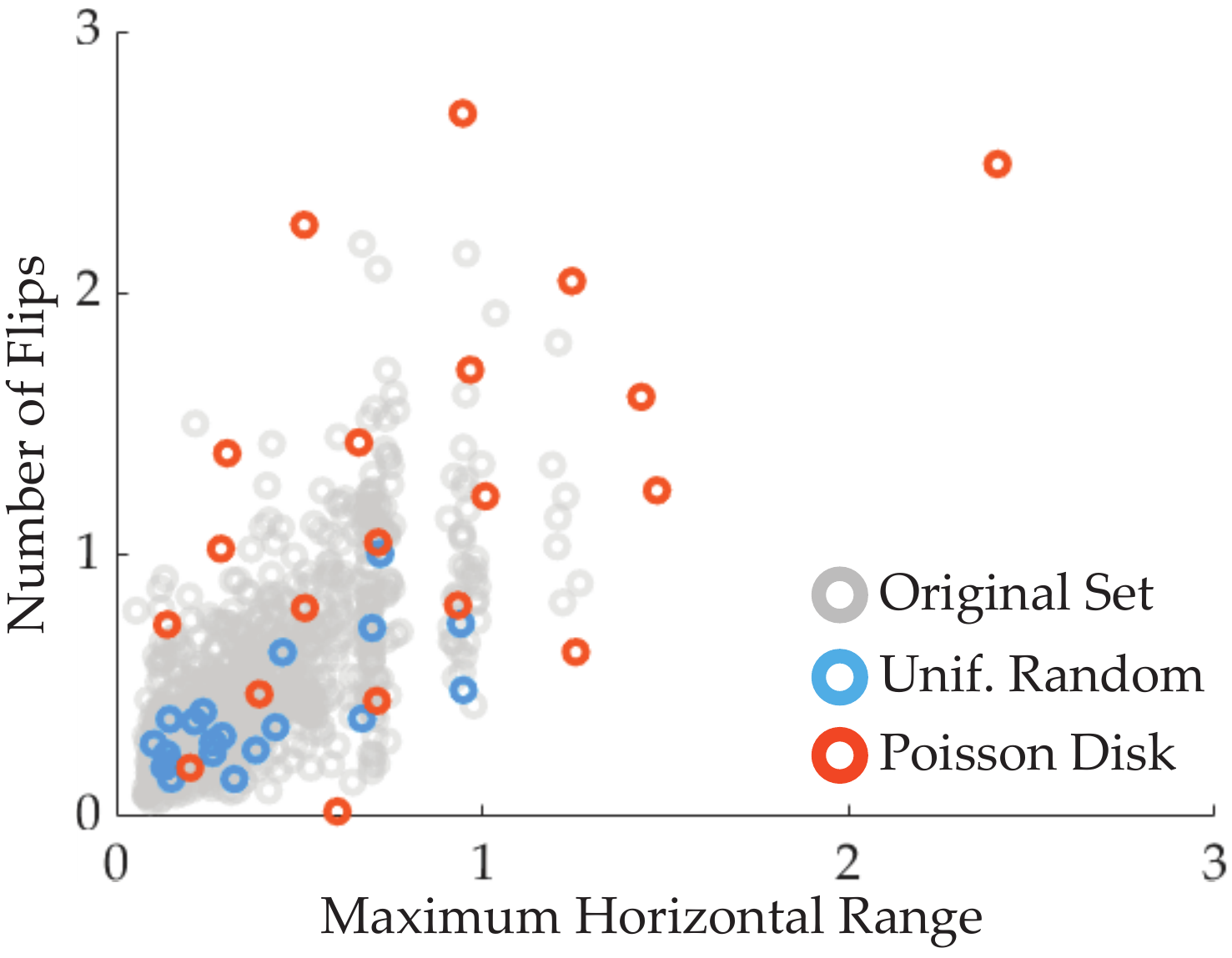}
	\centering
	\vspace{-10px}
	\caption{Features of $1000$ \emph{Tosser} trajectories are visualized in two-dimensional plane (gray). Poisson disk sampling allows us to obtain a diverse set of $20$ samples (orange), whereas sampling uniformly at random yields mostly uninteresting trajectories (blue).}
	\vspace{-20px}
	\label{fig:poisson}
\end{figure}

\vspace{5px}
\subsection{Baselines}
For our analyses, we compared three methods:
\begin{itemize}
	\item \textsc{RandomGP}: The reward is modeled using a Gaussian Process. The two distinct trajectories selected in each training query are sampled from a training dataset uniformly at random.
	\item \textsc{ActiveLinear}: The reward is modeled as a linear combination of features, and the active query generation method of \cite{biyik2019asking} selects the most informative comparison queries at every step of training.
	\item \textsc{ActiveGP}: The reward is modeled as a Gaussian Process. We will use our active query generation method to generate the most informative comparison queries to efficiently learn the reward function. 
\end{itemize}

We generated a training dataset of trajectories with uniformly randomly selected actions. At every iteration of \textsc{ActiveGP} and \textsc{ActiveLinear}, we computed the expected information gain of each possible query from this dataset to select the most informative query. This approach decreases the computation time compared to solving a continuous optimization over all possible trajectories as it was done by \cite{sadigh2017active, palan2019learning}.

\begin{figure*}[h]
	\includegraphics[width=\textwidth]{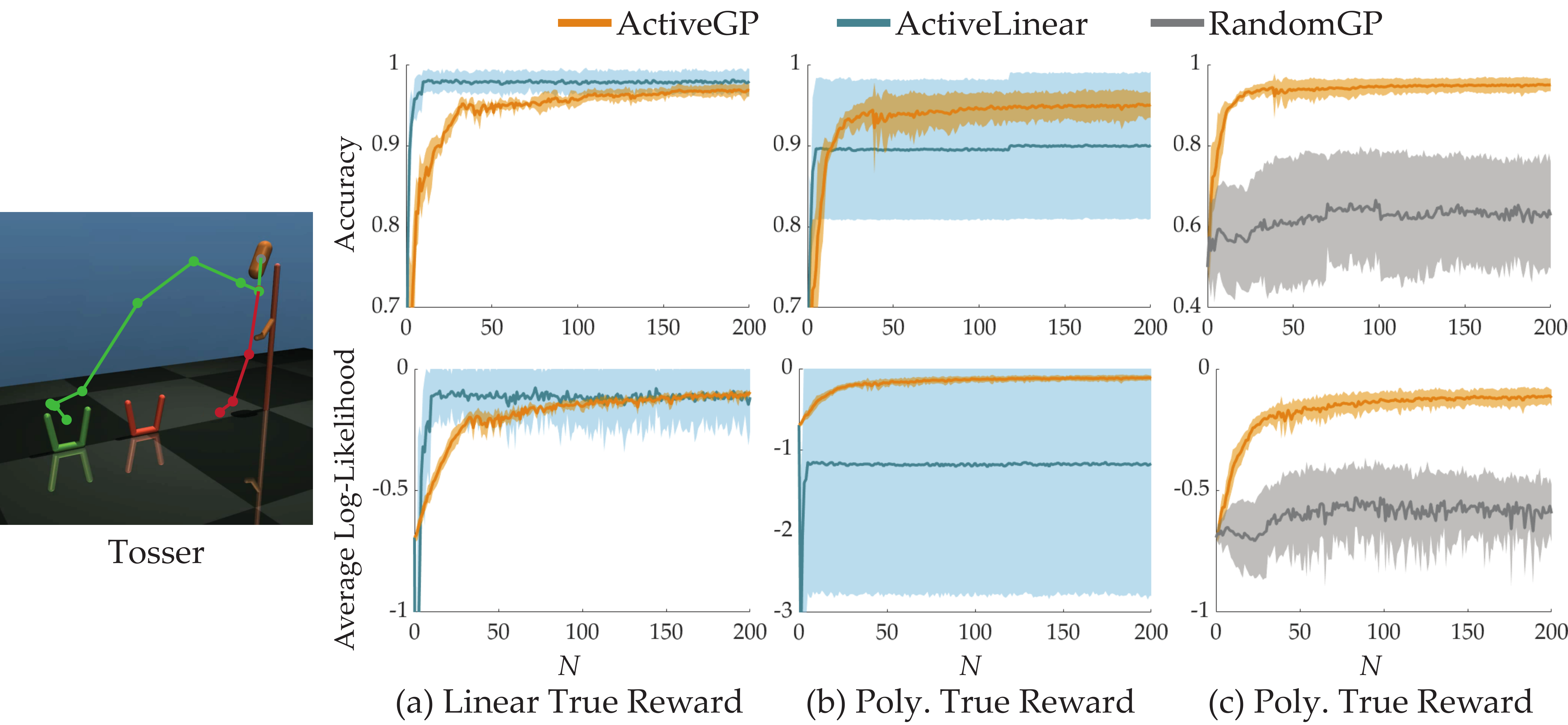}
	\centering
	\vspace{-20px}
	\caption{Accuracies and average log-likelihoods for test set queries are shown for the \emph{Tosser} environment (mean$\pm$std over $5$ runs). \textbf{(a)} Expressiveness results when the true underlying reward function is linear. \textbf{(b)} Expressiveness results when the true underlying reward function is a degree-of-two polynomial. \textbf{(c)} Data-efficiency results that compare \textsc{ActiveGP} with \textsc{RandomGP}. Accuracies and average log-likelihoods for test set queries are shown (mean$\pm$std). Active query generation improves data-efficiency over random querying in both tasks. This can be seen through both accuracy and log-likelihood.}
	\vspace{-15px}
	\label{fig:tosser_results}
\end{figure*}

\vspace{5px}
\subsection{Evaluation}
We compare GP reward with linear reward in terms of \emph{expressiveness} (\textsc{ActiveGP} vs. \textsc{ActiveLinear}), and compare active query generation with random querying baseline in terms of \emph{data-efficiency} (\textsc{ActiveGP} vs. \textsc{RandomGP}).

\vspace{5pt}
\noindent \textbf{Test Set Generation.}
For both analyses on the expressiveness and data-efficiency, we also generated test sets of trajectories from the same distribution as the training set. However, it would not be fair to use the test set as is. Obtained with uniformly random action sequences, the majority of the training set is uninteresting trajectories, e.g. the ego car moves slightly forward and backward (similar to a random walk) in \emph{Driver}, or the robot does not hit the capsule in \emph{Tosser}. Using the test set without further modifications would mean we give more importance to these uninteresting behaviors as they form the majority in the datasets. Obviously, this is not the case. We want to learn the reward function everywhere in the dynamically feasible region with equal importance. 

Hence, we adopted Poisson disk sampling \cite{bridson2007fast} to get a diverse set of trajectories from the test set. Poisson disk sampling makes sure the difference between trajectories\footnote{We used $L_2$ distance between the feature vectors.} is above some threshold by rejecting the samples that violate this constraint. A small example set of samples is compared to uniformly random samples in Fig.~\ref{fig:poisson} for the \emph{Tosser} environment.

After obtaining the diverse test set, we stored the true (noiseless) response of the simulated user for each possible query in this set. For the analysis on expressiveness, we computed the accuracy and the log-likelihood of the true responses under the reward functions that are learned with $N$ actively chosen queries (up to $N=200$). For data-efficiency analysis, we again used the true human responses to the queries in the diverse test set (only from the polynomial reward functions) to calculate the accuracy and the log-likelihood under the learned reward functions.

\noindent \textbf{Expressiveness.} Figs.~\ref{fig:driver_results}(a,b) and \ref{fig:tosser_results}(a,b) show the results of expressiveness simulations. When the true reward is polynomial, the linear model results in very high variance in both accuracy and likelihood, because its performance relies on how good a linear function can explain the true nonlinear reward. In this case, the GP model captures nonlinearities better than the linear model and provides better learning (Figs.~\ref{fig:driver_results}(b) and \ref{fig:tosser_results}(b)). When the true reward function is linear in features, a linear model naturally learns faster. However, as shown in Figs.~\ref{fig:driver_results}(a) and \ref{fig:tosser_results}(a), even in that case, GP model can achieve linear model's performance. To further improve the reward model, one can consider an approach to combine the linear and GP models by keeping a belief distribution over whether the true reward is linear or not, and actively querying the user according to this belief. We leave this extension as future work.

\vspace{5px}
\noindent \textbf{Data-Efficiency.} We then evaluated how our active query generation helps with data-efficiency. Figs.~\ref{fig:driver_results}(c) and \ref{fig:tosser_results}(c) compare \textsc{ActiveGP} and \textsc{RandomGP} for the simulation environments. It can be seen that active querying significantly accelerates learning over random querying. It should be noted that the number of samples taken via Poisson disk sampling matters: While choosing a very small number will increase the variance in the results, choosing a very large number will make random querying seem like it performs comparable to (or even better than) the active querying as the test set will mostly consist of uninteresting trajectories, which are also abundant in the training set, as we stated earlier.

\begin{figure}[t]
	\includegraphics[width=0.9\columnwidth]{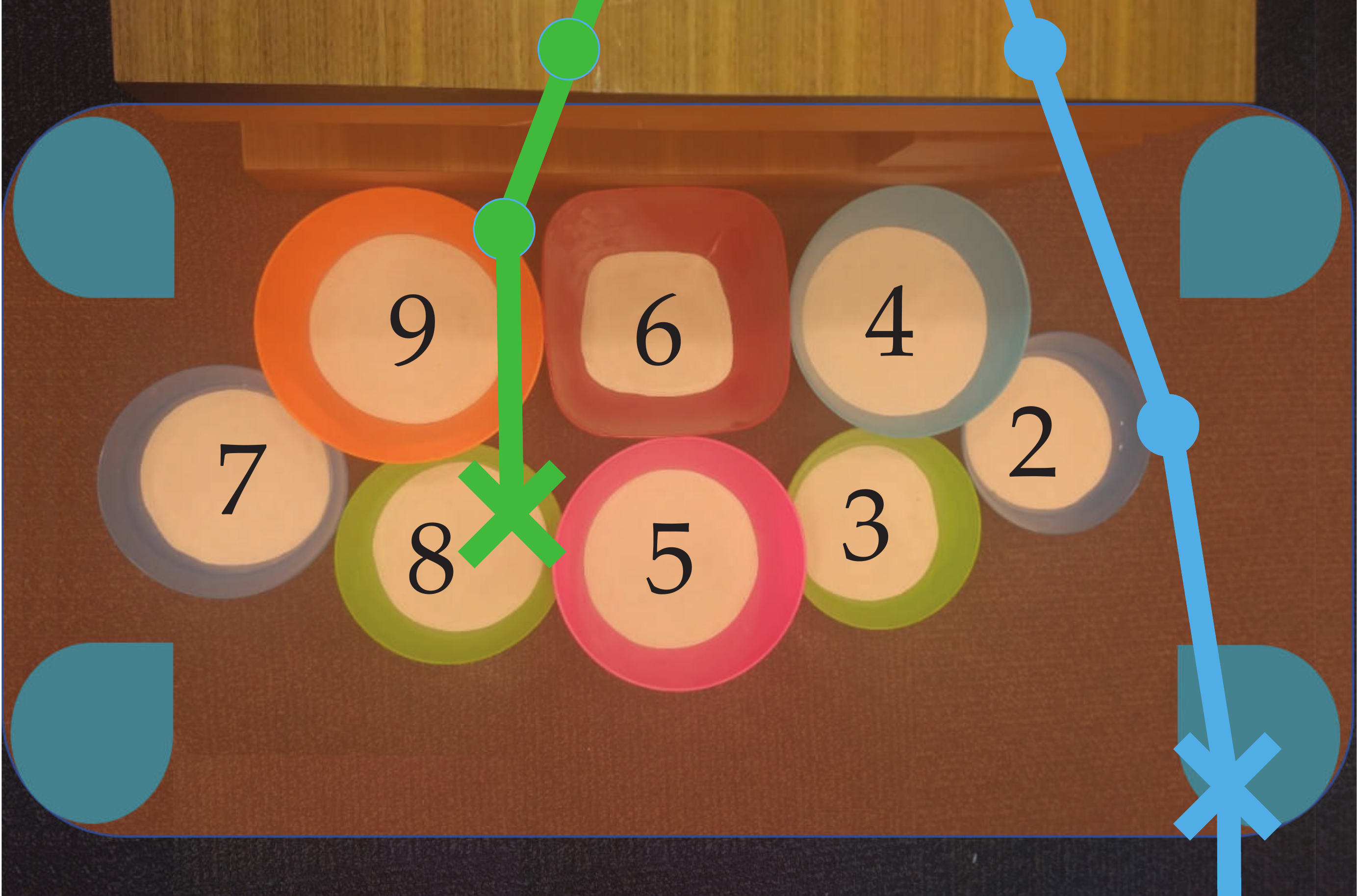}
	\centering
	\vspace{-7px}
	\caption{Top view of the eight targets in the variant of mini-golf user study. The users assign distinct scores from $2$ to $9$ to the targets. The figure shows an example of this ranking. While the robot is capable of hitting the ball into the entire shaded region, the maximizers of a linear reward always lie near the corners of the shaded region in blue. Therefore, while the GP reward model can query the user with better trajectories (e.g. the green trajectory), the linear model only explores the boundaries (e.g. the blue trajectory that throws the ball outside of this region). Crosses show where the ball hits the ground.
	}
	\label{fig:targets}
	\vspace{-16px}
\end{figure}

\section{User Studies}
\subsection{Experiment Setup}
\begin{figure*}[th]
	\includegraphics[width=\textwidth]{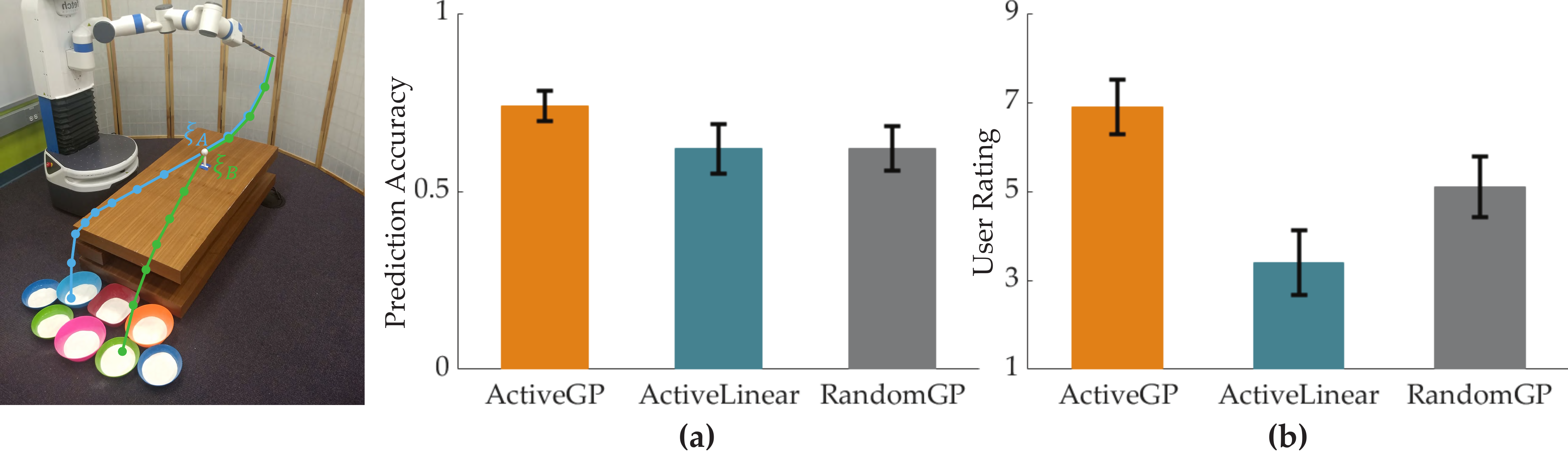}
	\centering
	\vspace{-20px}
	\caption{\textbf{(a)} Prediction accuracy results (mean$\pm$se). Each trained with $15$ queries, \textsc{ActiveGP} achieves significantly higher prediction accuracy than both \textsc{ActiveLinear} and \textsc{RandomGP} ($p<0.05$). \textbf{(b)} User ratings on the final robot performance (mean$\pm$se). \textsc{ActiveGP} accomplishes the task significantly better than both \textsc{ActiveLinear} and \textsc{RandomGP} ($p<0.05$).}
	\label{fig:fetch_results}
	\vspace{-15px}
\end{figure*}
We also compare our method \textsc{ActiveGP} with \textsc{ActiveLinear} and \textsc{RandomGP} on a user study with a Fetch mobile manipulator robot \cite{wise2016fetch}. In this study, the human subjects teach the Fetch robot how to play a variant of mini golf where the robot can achieve different scores by hitting the ball to different targets (see Fig.~\ref{fig:front_fig} and Fig.~\ref{fig:targets} for the set up). However, these scores are only known to the human. In fact, the robot does not even know the locations of the targets, and it tries to learn the reward as a function of its control inputs. Fixing some of the joints, we let the robot vary only its shot speed and angle, which are also the features of the reward function.

This experiment setting is interesting because a linear reward function can only encode whether the robot must hit the ball to the right or to the left, or whether it must hit with high or low speed. It cannot particularly encourage (or discourage) hitting with a modest angle and/or speed. Therefore, as we show in Fig.~\ref{fig:targets}, the targets that are around the middle region cannot be the maximizers of a linear reward function.

\vspace{5px}
\subsection{Subjects and Procedure}
We recruited 10 users (6 males, 4 females) with an age range from $19$ to $28$. Each user first assigned their distinct scores (from $2$ to $9$) to the eight targets. The robot then queried them with $50$ pairwise comparison questions: $15$ for \textsc{ActiveGP}, $15$ for \textsc{ActiveLinear}, $15$ for \textsc{RandomGP} and $5$ queries generated uniformly at random to create a test set. We shuffled the order of queries to avoid any bias. We used the reward models, each of which is learned with $15$ queries, to predict the user responses in the test set. The prediction score on the test set provides an accuracy metric.

In addition to the accuracy, we assessed whether the robot could successfully learn how to perform a good shot. For this, after the subjects responded to $50$ queries, the robot demonstrated $3$ more trajectories each of which correspond to the optimal trajectory of one method, the trajectory that maximizes the learned reward function. Again, the order of these trajectories was shuffled. After watching each demonstration, the subjects assigned a score to the shot from a 9-point Likert scale ($1$-very bad, $9$-very good).

\vspace{5px}
\subsection{Results and Discussion}
We provide a video that gives an overview of user studies and their results at \url{https://youtu.be/SLSO2lBj9Mw}.

Fig.~\ref{fig:fetch_results}(a) shows the prediction accuracy values on the test sets collected from the subjects (averaged over the subjects). By modeling the reward using a GP and querying the users with the most informative questions, \textsc{ActiveGP} achieves significantly higher prediction accuracy ($0.74\pm0.04$, mean$\pm$se) compared to both \textsc{ActiveLinear} ($0.62\pm0.07$) and \textsc{RandomGP} ($0.62\pm0.06$) with $p<0.05$ (Wilcoxon signed rank test). The results from this user study are aligned with our simulation user studies.

In reward learning, it is crucial to validate whether the learned reward function can encode the desired behavior or not. Fig.~\ref{fig:fetch_results}(b) shows the user ratings to the trajectories that the robot showed after learning the user preferences via $3$ different methods. \textsc{ActiveGP} obtains significantly higher scores ($6.9\pm0.6$) than both \textsc{ActiveLinear} ($3.4\pm0.7$) and \textsc{RandomGP} ($5.1\pm0.7$) with $p<0.05$. While \textsc{ActiveLinear} occasionally achieves high scores when the users' preferred target is near the edge, it generally fails to produce the desired behavior due to its low expressive power.

\section{Conclusion} \label{sec:conclusion}
\noindent\textbf{Summary.} We developed an active preference-based GP regression technique for reward learning. Our work tackles the lack of expressiveness of reward functions, data-inefficiency, and the incapability to demonstrate or quantify trajectories. Our results in simulations and user studies suggest our method is more successful in expressiveness and data-efficiency compared to the baselines.

\vspace{5px}
\noindent\textbf{Limitations and Future Work.} We developed our methods only for pairwise comparisons. While extending them to learning from rankings is not mathematically very complicated, its data-efficiency compared to pairwise comparisons needs thorough analysis. Similarly, one could easily incorporate options to denote user uncertainty, which was shown to ease the process for humans \cite{biyik2019asking}. GP regression becomes computationally heavy when the domain is high-dimensional (when $d$ is large). This is a limitation of our work due to the use of GPs, and can be alleviated through efficient rank-one GP update approximations. Finally, although our methods ease the feature design, there still needs to be a design stage\textemdash it is often unrealistic to hope $\Psi(\xi)=\xi$ will work, due to high dimensionality of $\Xi$. Further research is warranted to simultaneously learn both the reward function $f$ and the feature function $\Psi$.

\section*{Acknowledgments}
We thank Farid Soroush for the early discussions on active preference-based GP regression; Sydney M. Katz, Amir Maleki and Juan Carlos Aragon for the discussions on alternative ways to ease feature design. We acknowledge funding by Allstate, FLI grant RFP2-000, and NSF grants \#1941722 and \#1849952.


\bibliographystyle{unsrtnat}
\bibliography{refs}

\clearpage

\onecolumn

	\section{Appendix}
	\subsection{Active Query Generation Derivation}
	Let $\Sigma$ be the posterior covariance matrix between $f(\Psi^{(1)})$ and $f(\Psi^{(2)})$. And let
	\begin{align*}
	\Sigma^{-1} = \begin{bmatrix} c&d \\ d&c'  \end{bmatrix}.
	\end{align*}
	Throughout the derivation, all integrals are calculated over $\mathbb{R}$, but we drop it to simplify the notation. We write the first entropy term in the optimization \eqref{eq:problem} as:
	\begin{align*}
	H(q \mid \Psi, \bm{\Psi}, \bm{q}) &= h\left(\int \int\Phi\left(\frac{f^{(1)}-f^{(2)}}{\sqrt{2}\sigma}\right)\mathcal{N}([f^{(1)},f^{(2)}]\mid[\mu^{(1)},\mu^{(2)}], \Sigma)df^{(2)}df^{(1)}\right)\\
	&= h\left(\frac{\sqrt{cc'-d^2}}{2\pi }\int \int \Phi\left(\frac{f^{(1)}-f^{(2)}}{\sqrt{2}\sigma}\right)e^{-\frac{1}{2}(c(f^{(1)}-\mu^{(1)})^2 + c'(f^{(2)}-\mu^{(2)})^2+2d(f^{(1)}-\mu^{(1)})(f^{(2)}-\mu^{(2)}))}df^{(2)}df^{(1)}\right)\\
	&= h\left(\frac{\sqrt{cc'-d^2}}{2\pi}\int \int \Phi\left(\frac{f^{(1)}-f^{(2)}}{\sqrt{2}\sigma}\right)e^{-\frac{1}{2}(c((f^{(1)}-\mu^{(1)})^2+\frac{2d}{c}(f^{(1)}-\mu^{(1)})(f^{(2)}-\mu^{(2)}))+c'(f^{(2)}-\mu^{(2)})^2)}df^{(1)}df^{(2)}\right)\\
	&= h\left(\frac{\sqrt{cc'-d^2}}{2\pi}\int \int \Phi\left(\frac{f^{(1)}-f^{(2)}}{\sqrt{2}\sigma}\right)e^{-\frac{1}{2}(c(f^{(1)}-\mu^{(1)}+\frac{d}{c}(f^{(2)}-\mu^{(2)}))^2-\frac{d^2}{c}(f^{(2)}-\mu^{(2)})^2+c'(f^{(2)}-\mu^{(2)})^2)}df^{(1)}df^{(2)}\right)\\
	&= h\left(\frac{\sqrt{cc'-d^2}}{2\pi}\int e^{-\frac{1}{2}c'(f^{(2)}-\mu^{(2)})^2}e^{\frac{1}{2}\frac{d^2}{c}(f^{(2)}-\mu^{(2)})^2} \int \Phi\left(\frac{f^{(1)}-f^{(2)}}{\sqrt{2}\sigma}\right)e^{-\frac{1}{2}(c(f^{(1)}-\mu^{(1)}+\frac{d}{c}(f^{(2)}-\mu^{(2)}))^2)}df^{(1)}df^{(2)}\right)\\
	&= h\left(\frac{\sqrt{cc'-d^2}}{2\pi}\int e^{-\frac{1}{2}\frac{c'c-d^2}{c}(f^{(2)}-\mu^{(2)})^2} \int \frac{\Phi\left(\frac{f^{(1)}-f^{(2)}}{\sqrt{2}\sigma}\right)e^{-\frac{1}{2}(c(f^{(1)}-\mu^{(1)}+\frac{d}{c}(f^{(2)}-\mu^{(2)}))^2)}}{\frac{\sqrt{2\pi}}{\sqrt c}} \frac{\sqrt{2\pi}}{\sqrt c}df^{(1)}df^{(2)}\right)\\
	&= h\left(\frac{\sqrt{cc'-d^2}}{\sqrt{2\pi}\sqrt{c}}\int e^{-\frac{1}{2}\frac{c'c-d^2}{c}(f^{(2)}-\mu^{(2)})^2} \int \frac{\Phi\left(\frac{f^{(1)}-f^{(2)}}{\sqrt{2}\sigma}\right)e^{-\frac{1}{2}(c(f^{(1)}-\mu^{(1)}+\frac{d}{c}(f^{(2)}-\mu^{(2)}))^2)}}{\frac{\sqrt{ 2\pi}}{\sqrt c}}df^{(1)}df^{(2)}\right)
	\end{align*}
	Using the mathematical identity $\int_{x} \phi(x) N(x|\mu, \sigma^2) dx = \phi(\frac{\mu}{\sqrt{1+\sigma^2}})$, we obtain
	\begin{align*}
	H(q \mid \Psi, \bm{\Psi}, \bm{q}) &= h\left(\frac{\sqrt{cc'-d^2}}{\sqrt{2\pi}\sqrt{c}}\int e^{-\frac{1}{2}\frac{c'c-d^2}{c}(f^{(2)}-\mu^{(2)})^2} \Phi\left(\frac{\mu^{(1)}-\frac{d}{c}f^{(2)}+\frac{d}{c}\mu^{(2)}-f^{(2)}}{\sqrt{2}\sigma\sqrt{1+\frac{1}{2c\sigma^2}}}\right)df^{(2)}\right)\\
	&= h\left(\frac{\sqrt{cc'-d^2}}{\sqrt{2\pi}\sqrt{c}}\int e^{-\frac{1}{2}\frac{c'c-d^2}{c}(f^{(2)}-\mu^{(2)})^2} \Phi\left(\frac{\mu^{(1)}+\frac{d}{c}\mu^{(2)}-(\frac{d}{c}+1)f^{(2)}}{\sqrt{2}\sigma\sqrt{1+\frac{1}{2c\sigma^2}}}\right)df^{(2)}\right)\\
	&= h\left(\frac{\sqrt{cc'-d^2}}{\sqrt{2\pi}\sqrt{c}}\int e^{-\frac{1}{2}\frac{c'c-d^2}{c}(f^{(2)}-\mu^{(2)})^2} \frac{\Phi\left(\frac{-(\frac{d}{c}+1)(f^{(2)}-\frac{\mu^{(1)}+\frac{d}{c}\mu^{(2)}}{\frac{d}{c}+1})}{\sqrt{2}\sigma\sqrt{1+\frac{1}{2c\sigma^2}}}\right)}{\frac{\sqrt{2\pi c}}{\sqrt{c'c-d^2}}} \frac{\sqrt{2\pi c}}{\sqrt{c'c-d^2}}df^{(2)}\right)
	\end{align*}
	Using the same identity again,
	\begin{align*}
	H(q \mid \Psi, \bm{\Psi}, \bm{q}) &= h\left(\Phi\left(\frac{\mu^{(1)}-\mu^{(2)}}{\sqrt{2}\sigma \sqrt{1+\frac{1}{2c\sigma^2}}\sqrt{1+\frac{c}{2\sigma^2+\frac{1}{c}}\frac{(1+\frac{d}{c})^2}{c'c-d^2}}}\right)\right)
	\end{align*}
	One can then expand the expression in the denominator and use the facts that $\mathrm{Var}(f(\Psi^{(1)}))=\frac{c'}{cc'-d^2}$, $\mathrm{Var}(f(\Psi^{(2)}))=\frac{c}{cc'-d^2}$ and $\mathrm{Cov}(f(\Psi^{(1)}),f(\Psi^{(2)}))=\frac{-d}{cc'-d^2}$ to obtain
	\begin{align*}
	H(q \mid \Psi, \bm{\Psi}, \bm{q})=h\left(\Phi\left(\frac{\mu^{(1)}-\mu^{(2)}}{\sqrt{2\sigma^2 + g(\Psi^{(1)},\Psi^{(2)})}}\right)\right).
	\end{align*}
	where $g(\Psi^{(1)},\Psi^{(2)}) = \mathrm{Var}(f(\Psi^{(1)})) + \mathrm{Var}(f(\Psi^{(2)})) - 2\mathrm{Cov}(f(\Psi^{(1)}),f(\Psi^{(2)}))$
	
We next make the derivation for the second entropy term. To simplify the notation, we let $\sigma'^2=\frac{\pi ln(2)}{2}, \sigma''^2 = \sigma'^2 +\frac{1}{c}$, and $\sigma_b^2 = \frac{c(1+\frac{d}{c})^2}{c'c-d^2}$. By performing a linearization over the logarithm of the second entropy term as in \cite{houlsby2011bayesian},
	\begin{align*}
	\mathbb{E}_{f\sim P(f \mid \bm{\Psi}, \bm{q})}\left[H(q \mid \Psi, f)\right] &\approx \frac{\sqrt{c'c-d^2}}{2\pi}\int \int e^{-\frac{(f^{(1)}-f^{(2)})^2}{\pi ln(2)}}e^{-\frac{1}{2}(c(f^{(1)}-\mu^{(1)})^2 + c'(f^{(2)}-\mu^{(2)})^2+2d(f^{(1)}-\mu^{(1)})(f^{(2)}-\mu^{(2)}))}df^{(1)}df^{(2)}\\
	&= \frac{\sqrt{c'c-d^2}}{2\pi}\int e^{-\frac{1}{2}c'(f^{(2)}-\mu^{(2)})^2} \int e^{-\frac{(f^{(1)}-f^{(2)})^2}{2\sigma'^2 }}e^{-\frac{1}{2}(c(f^{(1)}-\mu^{(1)})^2 +2d(f^{(1)}-\mu^{(1)})(f^{(2)}-\mu^{(2)}))}df^{(1)}df^{(2)}\\
	&=\frac{\sqrt{c'c-d^2}}{2\pi}\int e^{-\frac{1}{2}c'{f^{(2)}}^2} \int e^{-\frac{(f^{(1)}+\mu^{(1)}-f^{(2)}-\mu^{(2)})^2}{2\sigma'^2}}e^{-\frac{1}{2}c{f^{(1)}}^2 -d f^{(1)} f^{(2)}}df^{(1)}df^{(2)}\\
	&=\frac{\sqrt{c'c-d^2}}{2\pi}\int e^{-\frac{1}{2}c'{f^{(2)}}^2} \int e^{-\frac{(f^{(1)}+\mu^{(1)}-f^{(2)}-\mu^{(2)})^2}{2\sigma'^2}}e^{-\frac{1}{2}c(f^{(1)}+\frac{d}{c}f^{(2)})^2+\frac{1}{2}\frac{d^2}{c}{f^{(2)}}^2}df^{(1)}df^{(2)}\\
	&=\frac{\sqrt{c'c-d^2}}{2\pi}\int e^{-\frac{1}{2}\frac{c'c-d^2}{c}{f^{(2)}}^2} \int e^{-\frac{(f^{(1)}-f^{(2)}+\mu^{(1)}-\mu^{(2)})^2}{2\sigma'^2}}e^{-\frac{1}{2}c(f^{(1)}+\frac{d}{c}f^{(2)})^2}df^{(1)}df^{(2)}
	\end{align*}
	By the change of variables for the inner integral with $u=f^{(1)} + \frac{d}{c}f^{(2)}$,
	\begin{align*}
	\mathbb{E}_{f\sim P(f \mid \bm{\Psi}, \bm{q})}\left[H(q \mid \Psi, f)\right] &=\frac{\sqrt{c'c-d^2}}{2\pi}\int e^{-\frac{1}{2}\frac{c'c-d^2}{c}{f^{(2)}}^2} \int_u e^{-\frac{(u-\frac{d}{c}f^{(2)}+\mu^{(1)}-f^{(2)}-\mu^{(2)})^2}{2\sigma'^2}}e^{-\frac{1}{2}cu^2}dudf^{(2)}\\
	&=\frac{\sqrt{c'c-d^2}}{2\pi}\int e^{-\frac{1}{2}\frac{c'c-d^2}{c}{f^{(2)}}^2} \int_u e^{-\frac{((1+\frac{d}{c})f^{(2)}-u-\mu^{(1)}+\mu^{(2)})^2}{2\sigma'^2}}e^{-\frac{1}{2}cu^2}dudf^{(2)}
	\end{align*}
	By another change of variables for the outer integral with $v=\frac{f^{(2)}}{1 + d/c}$,
	\begin{align*}
	\mathbb{E}_{f\sim P(f \mid \bm{\Psi}, \bm{q})}\left[H(q \mid \Psi, f)\right]&=\frac{1}{1+\frac{d}{c}}\frac{\sqrt{c'c-d^2}}{2\pi}\int_v e^{-\frac{1}{2}\frac{c'c-d^2}{c}\frac{v^2}{(1+\frac{d}{c})^2}} \int_u e^{-\frac{(v-u+\mu^{(2)}-\mu^{(1)})^2}{2\sigma'^2}}e^{-\frac{1}{2}cu^2}dudv
	\end{align*}
	By identifying the inner integral as a convolution of two Gaussians, we get
	\begin{align*}
	\mathbb{E}_{f\sim P(f \mid \bm{\Psi}, \bm{q})}\left[H(q \mid \Psi, f)\right] &=\frac{1}{1+\frac{d}{c}}\frac{\sqrt{c'c-d^2}}{2\pi}2\pi\sigma'\frac{1}{\sqrt{c}}\int_v e^{-\frac{1}{2}\frac{c'c-d^2}{c}\frac{v^2}{(1+\frac{d}{c})^2}} \frac{1}{\sqrt{2\pi}\sqrt{\sigma'^2+\frac{1}{c}}} e^{-\frac{1}{2}\frac{(v-(\mu^{(1)}-\mu^{(2)}))^2}{\sigma'^2+\frac{1}{c}}}dv\\
	&=\frac{1}{1+\frac{d}{c}}\sqrt{c'c-d^2}\sigma'\frac{1}{\sqrt{c}}\frac{1}{\sqrt{2\pi}\sqrt{\sigma'^2+\frac{1}{c}}}\int_v e^{-\frac{1}{2}\frac{v^2}{\sigma_b^2}}  e^{-\frac{1}{2}\frac{(v-(\mu^{(1)}-\mu^{(2)}))^2}{\sigma''^2}}dv
	\end{align*}
	By repeating the same convolution trick for the second integral,
	\begin{align*}
	\mathbb{E}_{f\sim P(f \mid \bm{\Psi}, \bm{q})}\left[H(q \mid \Psi, f)\right]
	&=\frac{1}{1+\frac{d}{c}}\sqrt{c'c-d^2}\sigma'\frac{1}{\sqrt{c}}\frac{1}{\sqrt{2\pi}\sqrt{\sigma'^2+\frac{1}{c}}} 2\pi \sigma_b \sigma'' \frac{1}{\sqrt{2\pi}\sqrt{\sigma_b^2+\sigma''^2}}e^{-\frac{1}{2}\frac{(\mu^{(1)}-\mu^{(2)})^2}{\sigma_b^2+\sigma''^2}}\\
	&=\frac{1}{1+\frac{d}{c}}\sqrt{c'c-d^2}\sigma'\frac{1}{\sqrt{c}}\frac{1}{\sqrt{\sigma'^2+\frac{1}{c}}}  \sigma_b \sigma'' \frac{1}{\sqrt{\sigma_b^2+\sigma''^2}}e^{-\frac{1}{2}\frac{(\mu^{(1)}-\mu^{(2)})^2}{\sigma_b^2+\sigma''^2}}
	\end{align*}
	
	Again, we express this in terms of covariance and variance expressions:
	\begin{align*}
	\mathbb{E}_{f\sim P(f \mid \bm{\Psi}, \bm{q})}\left[H(q \mid \Psi, f)\right] = \frac{\sqrt{\pi\ln(2)\sigma^2}\exp\left(-\frac{(\mu_{a} - \mu{b})^2}{\pi\ln(2)\sigma^2 +2g(\Psi^{(1)},\Psi^{(2)})}\right)}{\sqrt{\pi\ln(2)\sigma^2 + 2g(\Psi^{(1)},\Psi^{(2)})}}
	\end{align*}

\end{document}